\def\year{2018}\relax
\documentclass[letterpaper]{article} %DO NOT CHGGE THIS
\usepackage{aaai18}  %Required
\usepackage{times}  %Required
\usepackage{helvet}  %Required
\usepackage{courier}  %Required
\usepackage{url}  %Required
\usepackage{graphicx}  %Required
\usepackage[ruled,vlined]{algorithm2e}
\begin{document}
% The file aaai.sty is the style file for AAAI Press 
% proceedings, working notes, and technical reports.
%
\title{ Deep Stereo Matching with Explicit Cost Aggregation Sub-Architecture}
\author{Lidong Yu\textsuperscript{1}, Yucheng Wang\textsuperscript{2}, Yuwei Wu\textsuperscript{1}\thanks{corresponding author}, and Yunde Jia\textsuperscript{1}\\
\textsuperscript{1}Beijing Laboratory of Intelligent Information Technology, School of Computer Science,\\
Beijing Institute of Technology, Beijing, 100081\\
\textsuperscript{2}Kandao Australia Research Center\\
Suite 3.05 1 Richardson Place North Ryde NSW 2113, Australia\\
\{yulidong, wuyuwei, jiayunde\}@bit.edu.cn, wyc@kandaovr.com
}
\maketitle
\begin{abstract}
Deep neural networks have shown excellent performance for stereo matching. Many efforts focus on the feature extraction and similarity measurement of the matching cost computation step while less attention is paid on cost aggregation which is crucial for stereo matching. In this paper, we present a learning-based cost aggregation method for stereo matching by a novel sub-architecture in the end-to-end trainable pipeline. We reformulate the cost aggregation as a learning process of the generation and selection of cost aggregation proposals which indicate the possible cost aggregation results. The cost aggregation sub-architecture is realized by a two-stream network: one for the generation of cost aggregation proposals, the other for the selection of the proposals. The criterion for the selection is determined by the low-level structure information obtained from a light convolutional network. The two-stream network offers a global view guidance for the cost aggregation to rectify the mismatching value stemming from the limited view of the matching cost computation. The comprehensive experiments on challenge datasets such as KITTI and Scene Flow show that our method outperforms the state-of-the-art methods.
\end{abstract}
\section{Introduction}
Stereo matching is one of the fundamental problems in computer vision community. The goal of stereo matching is to compute a disparity map from images collected by stereo cameras. The disparity map is widely used in 3D scene reconstruction, robotics, and autonomous driving. Driven by the emergence of large-scale data sets and fast development of computation power, deep neural networks have proven effective for stereo matching. Many state-of-the-art methods raise the performance by learning robust local features or similarity measurements for cost computation \cite{zbontar2015computing,luo2016efficient,shaked2016improved}. However, these methods still have difficulties in textureless areas and occluded regions because of the limited view field during cost computation.

\begin{figure}
\centering  
\includegraphics[width=8.3cm]{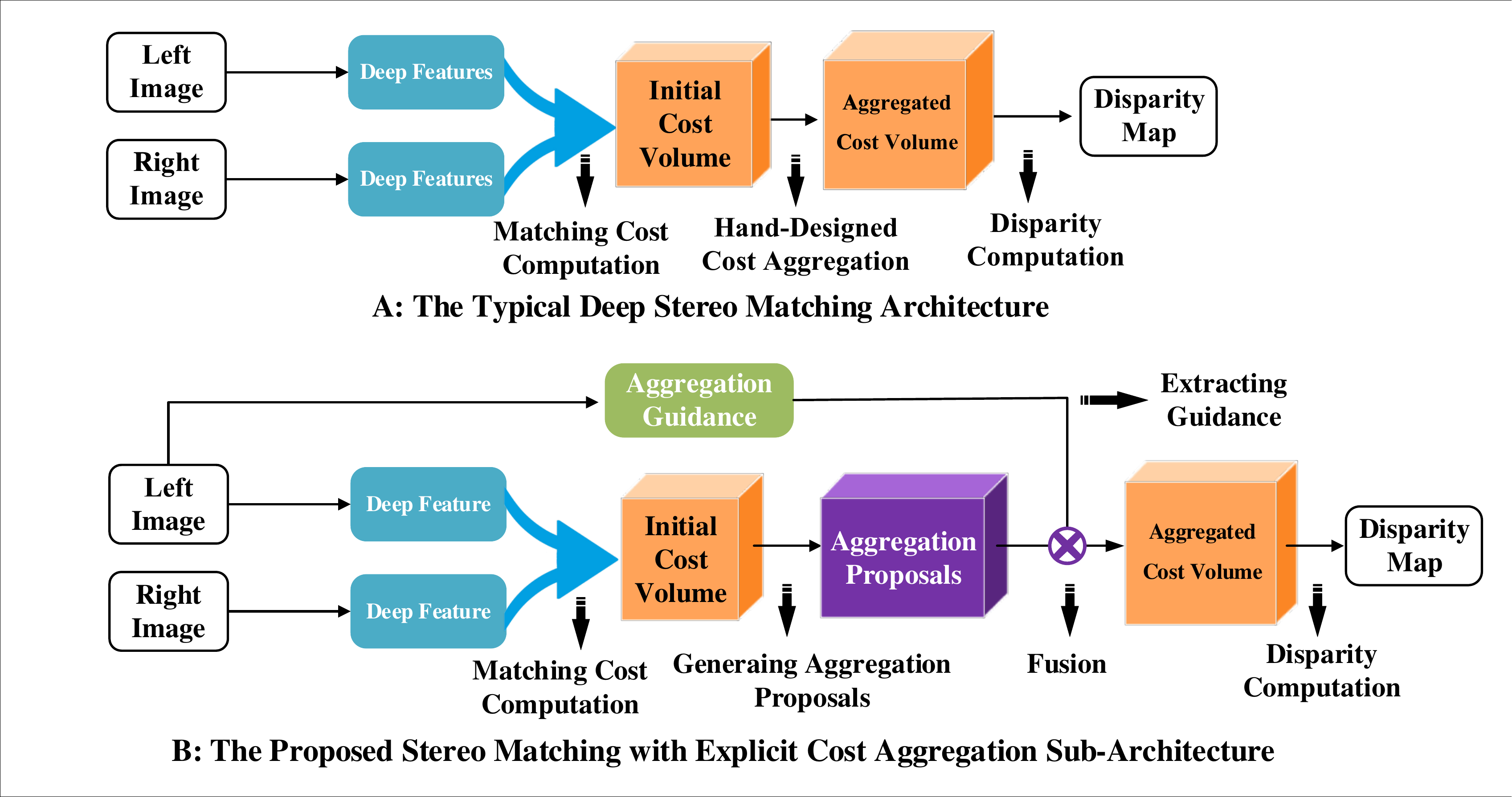}
\caption{Comparisons between the typical deep stereo matching pipeline and pipeline with our learning-based cost aggregation. The architecture A is the typical deep stereo matching pipeline with traditional cost aggregation method. The architecture B is our learning-based cost aggregation. The details of architectures will be shown in Figure \ref{fig:all}, where the parts are matching according to the colors.
}  
\label{fig:first} 
\end{figure}

To handle mismatching values of the cost computation results, which is called cost volume, the cost aggregation step is indispensable in traditional stereo matching methods. Cost aggregation is applied to the cost volume to rectify the incorrect values by aggregating the computed matching cost. It is typically performed by summing or averaging the matching cost over a support region within a constant disparity \cite{yang2012non,min2011revisit,tombari2008classification}. However, the traditional cost aggregation methods are limited by the shallow, hand-designed scheme to perform the aggregation. They cannot effectively take global view guidance into account while keeping the local fitness. In this paper, we propose a learning-based cost aggregation to keep the balance between global view and local fitness using a novel two-stream neural network.

\begin{figure*}
\centering  
\includegraphics[height=7cm]{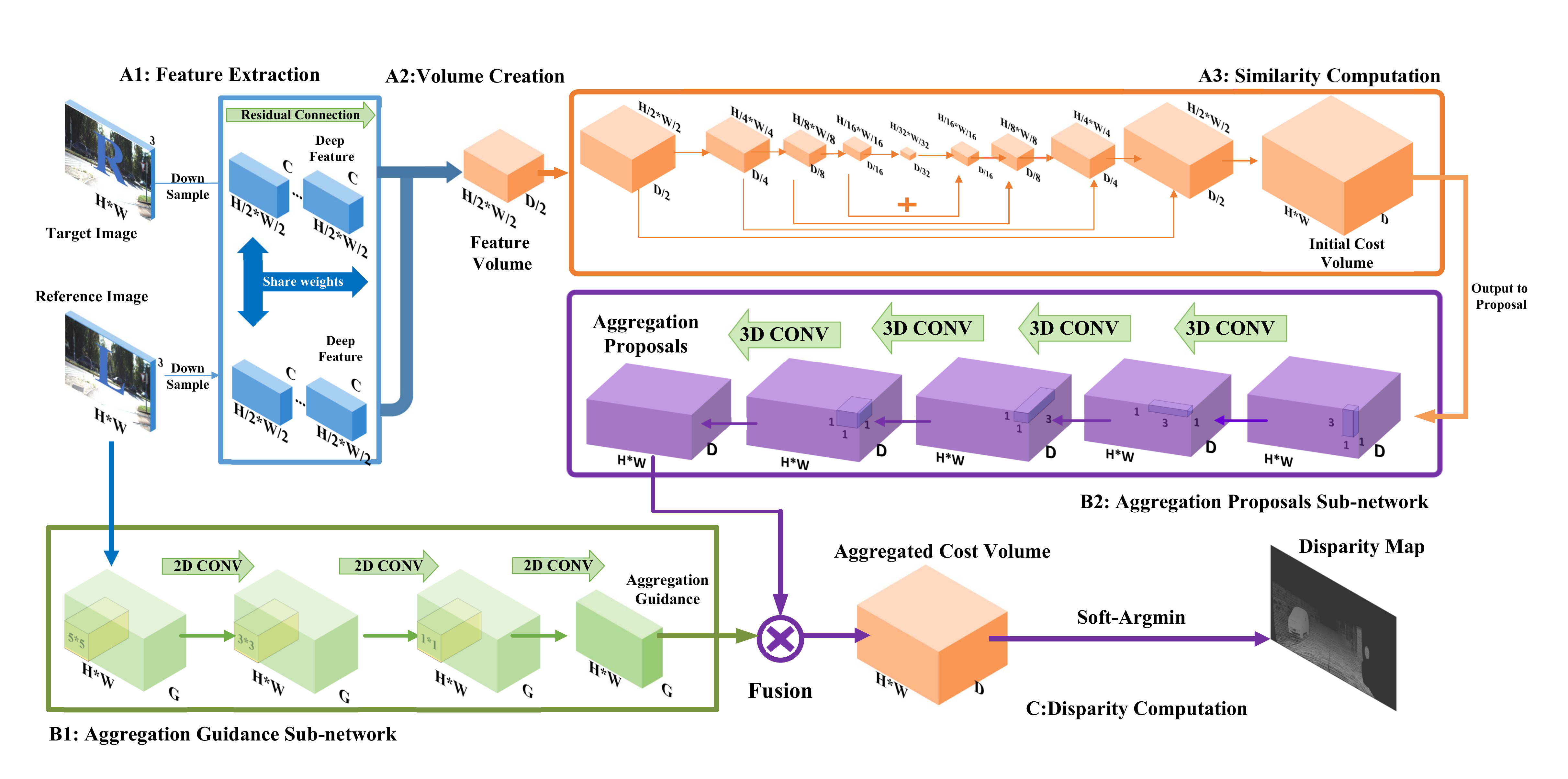}
\caption{Our stereo matching pipeline with learning-based cost aggregation. The different colors represent the different size of data: blue for $W\times H\times C$, orange for $D\times H\times W\times C$, green for $H\times W\times G$ and purple for $D\times H\times W\times G$. The cost computation step is divided into three components. A1 is a feature extraction sub-network using the residual Siamese network. A2 is a feature volume construction part, and the detailed illustration is shown in Figure \ref{fig:volume}. A3 computes the similarity between the feature volume using a 3D convolutional network and produces the initial cost volume. The learning-based cost aggregation is carried out by a two-stream network as a sub-architecture for the whole pipeline. The guidance stream is illustrated in B1. The proposals stream is shown in B2. The cost aggregation result is obtained by a winner-take-all strategy to select the best proposal. Finally, a soft-argmin function is employed to compute the disparity map.
} 
\label{fig:all} 
\end{figure*}

The proposed cost aggregation can incorporate with other deep stereo matching pipeline in an end-to-end manner because it is conducted as a sub-architecture for the whole network. With the learning-based cost aggregation, the end-to-end trainable stereo matching pipeline can not only learn the feature and similarity measurementment for cost computation but also perform the cost aggregation. The comparisons of the proposed architecture with typical deep stereo pipelines are shown in Figure \ref{fig:first}. We can see that the learning-based cost aggregation is carried out by a two-stream network in an explicit manner. 

The cost aggregation process is reformulated as a learning mechanism to generate potential cost aggregation results called proposals and select the best one. Accordingly, the learning-based cost aggregation is carried out by a two-stream network: one stream is used for generating the proposals and the other stream is employed for evaluating proposals. The first stream holds the local fitness by generating potential aggregation results according to the cost volume computed from the cost computation. The generation is performed by a convolutional operation along the three dimensions of the cost volume, which aggregates information both on the spatial and depth space. The second stream brings in global view guidance for the cost aggregation by evaluating each proposal. For each proposal, it is evaluated by the guidance with the same size of the image, which is considered as the global view guidance. The guidance is obtained by a light convolutional network to bring in low-level structure information which is treated as the evaluation criterion for proposals. Since the structure information only contains 2D information, which is independent in depth, the guidance is unchanged along the depth dimension. Therefore, the evaluation for each proposal shares the same guidance for different disparities. After evaluating each proposal, a winner-take-all strategy is employed to choose the best-aggregated value to form the aggregated cost volume

The proposed architecture reaches a promising accuracy on the Scene Flow \cite{mayer2016large} and the KITTI benchmark \cite{menze2015object,geiger2012we}. Our contributions are three-fold.
\begin{itemize}
\item This work is, to the best of our knowledge, the first to explicitly model the cost aggregation in a learning-based scheme for stereo matching. We reformulate the cost aggregation as the learning process of generation and selection of cost aggregation proposals.
\item We propose a novel two-stream network to carry out the generation and selection of cost aggregation proposals. The proposed two-stream network maintains the rich semantic information while brings in low-level structure information, which demonstrates the ability to fuse the high-level feature with the low-level feature.
\item The proposed learning-based cost aggregation is carried out as a sub-architecture of the deep stereo matching pipeline in an end-to-end trainable manner. It is flexible for the pipelines without cost aggregation to raise accuracy.
\end{itemize}

\section{Related Work}

\subsection{Deep neural networks for Cost computation}
Using deep neural networks for stereo matching was firstly introduced by Zbontar et al. \cite{zbontar2015computing} with a Siamese network for cost computation. Luo et al. \cite{luo2016efficient} reduced the computation time by replacing the full-connection layer with an inner product. For the stereo matching task, the Siamese network is responsible for extracting deep representations for each pixel. The original simple convolutional layers are limited to generate the rich semantic representation, so the improved highway network such as the residual network was employed to improve representations under the Siamese architecture \cite{shaked2016improved,XuCVPR2017DCFlow}. Then a similarity measurementment is applied to compute the matching cost between corresponding pixels. Inspired by the progress of the dense pixel-wise task such as optical flow and semantics segmentation, the 3D auto-encoder shows excellent performance by a large view field. The closely work with our method is GC-Net which is an end-to-end pipeline using a 3D auto-encoder as the similarity measurement \cite{kendall2017end}. Similarly, we utilize the residual Siamese network for feature extraction and leverage the 3D auto-encoder to compute the similarity. For deep stereo matching pipelines, the use of volume processing has been proven effective to combine the feature extraction and similarity measurement  \cite{XuCVPR2017DCFlow}. We modify the traditional concatenating construction with an additional shift operation to construct a more effective feature volume.  

Despite the usage of deep neural networks for cost computation improve the stereo matching performance, it still has limitations on textureless areas, weak structure, and occluded regions. Hand-designed cost aggregation methods are normally used on the initial cost volume, whose improvement is barely adequate \cite{zbontar2015computing,luo2016efficient}. In this paper, we present a learnable cost aggregation method which can collaborate with deep cost computation methods in an end-to-end trainable manner. The two-stream network is shown effective to fuse different classes of features in video action recognition \cite{simonyan2014two}. Inspired by this, we design a novel two-stream network to carry out the cost aggregation. The two-stream network is presented to maintain the rich semantics of the cost computation while bringing into low-level structure information to guide the cost aggregation. The low-level structure information can be used as the global view guidance by a light neural network architecture \cite{mahendran2015understanding,zeiler2014visualizing}. The fusion of two-stream network is always realized by a concatenating function \cite{feichtenhofer2016convolutional}, in contrast, we introduce a winner-take-all strategy to fuse the two streams.

\subsection{ Cost Aggregation}
According to the taxonomy of stereo matching \cite{scharstein2002taxonomy}, a typical stereo matching pipeline can be divided into four steps: matching cost computation, cost aggregation, disparity computation, and disparity refinement. Many cost aggregation methods have been proposed to obtain high-quality disparity maps. Normally, most of them were performed locally by aggregating the matching cost value among a support region within the same disparity \cite{min2011revisit}. The traditional cost aggregation is implemented by the construction of support regions obtained by a similarity function that can measurement the similarity between two potentially related pixels in the same reference image  \cite{yang2012non}. Yoon and Kweon et al. proposed an adaptive support region approach whose similarity function can be interpreted as a variant of joint bilateral filtering \cite{Yoon2006Adaptive}. Cross-based approaches use a shape-adaptive window which consists of multiple horizontal lines spanning adjacent vertical rows based on the function of the color similarity and an implicit connectivity constraint \cite{zhang2009cross}. A more thorough overview of cost aggregation methods can be found in \cite{min2011revisit}. Most traditional methods, however, are limited by the shallow, hand-designed similarity function which cannot adequately build the support region for the cost aggregation. The usage of deep neural networks for cost aggregation can collaborate with deep cost computation methods in a trainable manner.

With the superiority of the two-stream architecture \cite{simonyan2014two,feichtenhofer2016convolutional}, we propose an explicit learning-based cost aggregation. In this paper, we reformulate the cost aggregation process as the generation and selection of cost aggregation proposals. The proposals are obtained by generating potential cost aggregation results from the initial cost volume. The selection of proposals uses the structure information as global view guidance in a winner-take-all (WTA) strategy.

\section{Network Architecture}
As a pixel-wise matching task, stereo matching is required to compute similarities between each pixel in the left image with $D$ corresponding pixels in right image, where $D$ is the maximum disparity. The computed matching cost can form the cost volume $C_0 (h,w,d)$. The stereo matching pipeline with the proposed cost aggregation is carried out by an end-to-end trainable network. Compared with using networks as a black box, we take experience from classical stereo matching algorithm \cite{scharstein2002taxonomy} to conduct the cost aggregation explicitly by a two-stream network. In this paper, unless otherwise specified, we refer to the left image as the reference image and the right image as the target image, where the disparity is computed from the reference image. 
\begin{table}[tbp]
\small
\centering  % 表居中
\caption{Architecture for Feature Extraction}
\begin{tabular}{l|l|l}  % {lccc} 表示各列元素对齐方式,left-l,right-r,center-c
\hline
Index &layer&output \\ \hline \hline % \hline 在此行下面画一横线
1 &$5\times 5\times 32$ stride 2&$1/2H\times 1/2W\times F$ \\       % & 表示列的分隔线
2-17 &$3\times 3\times 32$ stride 2&$1/2H\times 1/2W\times F$ \\        % \\ 表示重新开始一行
&residual connection*8&\\
18 &$3\times 3\times 32$ stride 2&$1/2H\times 1/2W\times F$ \\     \hline   % & 表示列的分隔线
        % \\ 表示重新开始一行
\hline
\end{tabular}

\label{tab:feature} 
\end{table}

The overview of our method is illustrated in Figure \ref{fig:all}. The matching cost computation can be divided into three parts: feature extraction, volume construction and similarity computation, as shown in Figure \ref{fig:all}.A1, Figure \ref{fig:all}.A2 and Figure \ref{fig:all}.A3, respectively. The detailed volume construction method is elucidated in Figure \ref{fig:volume}. A two-stream network carries out the proposed learning-based cost aggregation: the proposal network and the guidance network which are illustrated in Figure \ref{fig:all}.B2 and Figure \ref{fig:all}.B1, respectively. The disparity computation is shown in Figure \ref{fig:all}.C, the detailed implementation of C will be discussed later in this section.

\subsection{Matching Cost Computation}
Matching cost computation is designed to compute the similarity between corresponding pixels at the reference image and the target image. The disparity map can then be obtained from the cost volume. To determine the pixel-wise matching cost, we firstly generate deep representations for each pixel using a residual Siamese network. Then outputs from the Siamese network is fed into the feature volume construction which can transform features into the volume. Finally, the similarity measurement using 3D auto-encoder is applied on the feature volume to compute the matching cost volume.
\subsubsection{A1: Feature Extraction}
To compute the similarity between two pixels, we require a powerful representation for each pixel. Compared with the traditional raw pixel intensities, deep feature representation is more effective and robust to mitigate textureless regions and thin structure. As shown in Figure \ref{fig:all}.A1, we describe a Siamese network to extract the feature of each pixel. The Siamese network consists of two shared-weight sub-networks which concurrently deal with two input images. Each sub-network is composed of several residual blocks each of which consists of two $3 \times 3$ convolutional layers. To reduce the computational demand, we apply a $5 \times 5$ convolutional layer with $2 \times 2$ stride as a sub-sampling operation before the residual connection. For each residual block, it is activated before the residual operation. Each convolutional layer is followed by a batch normalized layer and a rectified linear unit except the last layer. From the detailed layer setting shown in Table \ref{tab:feature}, we can see that the result of the Siamese network produces two $H/2 \times W/2 \times F$ feature maps, where $H$ and $W$ denotes original input images size and $F$ indicates the filter channel. The two feature maps contain the deep feature for each pixel in the reference image and the target image, respectively.
\begin{figure}
\centering  
\includegraphics[width=8cm,height=4cm]{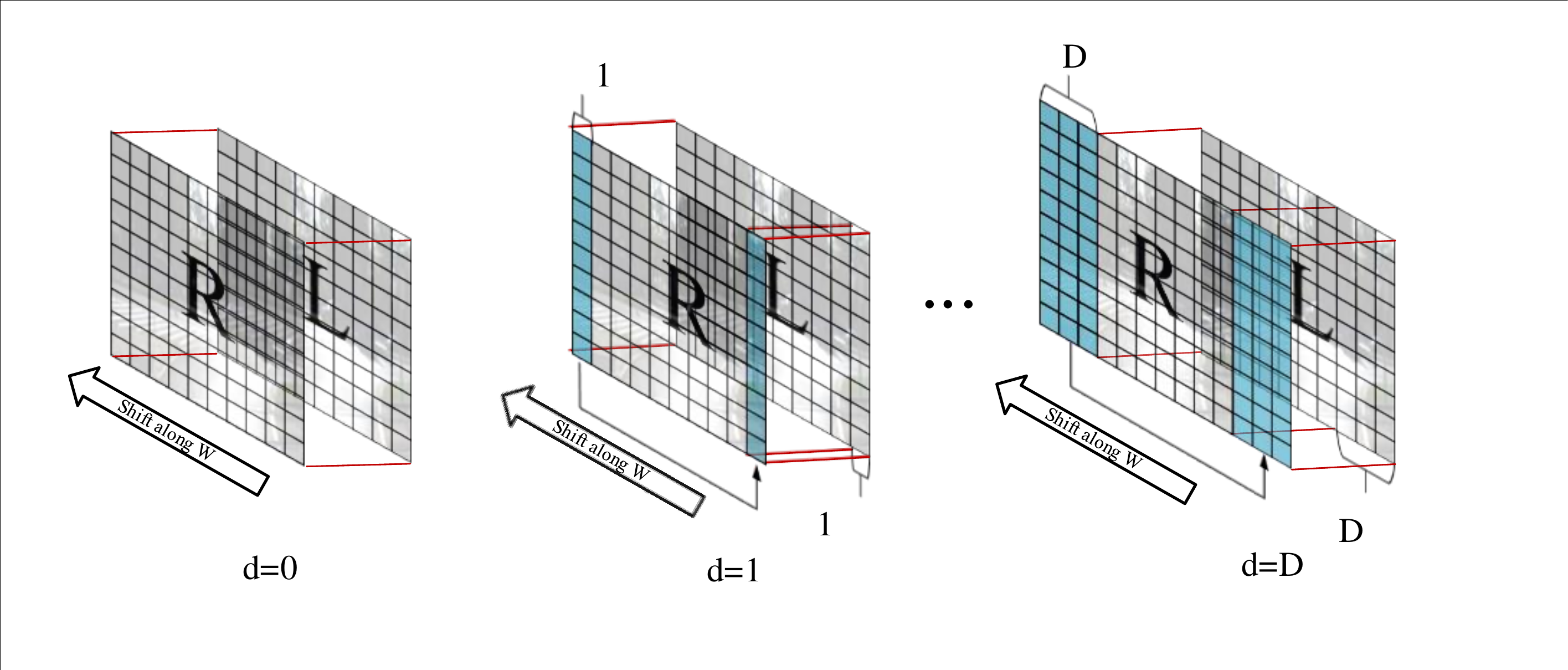}
\caption{The operation for the feature volume construction. Each grid square represents the feature for the pixel. We can simply employ a shift and concatenation operation to form a feature volume.}  
\label{fig:volume} 
\end{figure}
\subsubsection{A2: Feature Volume Construction}
Obtained the representation of each pixel, the next step is to compute the similarities between pixels. Since the volume input can be effective for the 3D convolutional computation, we transform the extracted features into a feature volume which contains the underlying group of pixels. Each element of the feature volume represents the feature for computation of the similarity between two pixels. Because input images have been rectified, we can simply employ a shift operation to form the feature volume. We set the output of the left sub-network as the base feature and the output from the right as the shift feature. The base feature is awaiting to be concatenating at the bottom, and the shift feature slides on the base feature. As depicted in Figure \ref{fig:volume}, the shift feature slides on base feature and concatenates with the base feature along feature channel. The mathematical definition is given by
\begin{equation}F(d,h,w)=B(h,w)\oplus S(d,h,(w+d)\bmod w),\end{equation}
where $B$ represents the base feature, $S$ denotes the shift feature and $\oplus$ indicates the concatenating operation. After packing the concatenating results, we get a 4D feature volume of $ D\times H/2\times W/2\times 2F $ size, where $D$ denotes the maximum disparity.

\begin{table}[tbp]
\scriptsize
\caption{Architecture for cost computation. Each layer except layer 37 is followed by batch normalization and ReLU. Layer 33-37 are 3D-deconvolutional layers.}
\centering  % 表居中
\begin{tabular}{l|c|c}  % {lccc} 表示各列元素对齐方式,left-l,right-r,center-c
\hline
Index &layer&output \\ \hline \hline % \hline 在此行下面画一横线
input &Volume Construction&$1/2D\times 1/2H\times 1/2W\times 2F$\\  \hline
19 &$3\times 3\times 3\times 32$ stride 1&$1/2D\times 1/2H\times 1/2W\times F$ \\       % & 表示列的分隔线
 \hline
20  &$3\times 3\times 3\times 32$ stride 1&$1/2D\times 1/2H\times 1/2W\times F$ \\       % & 表示列的分隔线
 \hline
 21  &$3\times 3\times 3\times 64$ stride 2&$1/4D\times 1/4H\times 1/4W\times 2F$ \\       % & 表示列的分隔线
 \hline
 22 &$3\times 3\times 3\times 64$ stride 1&$1/4D\times 1/4H\times 1/4W\times 2F$ \\       % & 表示列的分隔线
 \hline
 23 &$3\times 3\times 3\times 64$ stride 1&$1/4D\times 1/4H\times 1/4W\times 2F$ \\       % & 表示列的分隔线
 \hline
 24 &$3\times 3\times 3\times 64$ stride 2&$1/8D\times 1/8H\times 1/8W\times 2F$ \\       % & 表示列的分隔线
 \hline
 25 &$3\times 3\times 3\times 64$ stride 1&$1/8D\times 1/8H\times 1/8W\times 2F$ \\       % & 表示列的分隔线
 \hline
 26 &$3\times 3\times 3\times 64$ stride 1&$1/8D\times 1/8H\times 1/8W\times 2F$ \\       % & 表示列的分隔线
 \hline
 27 &$3\times 3\times 3\times 64$ stride 2&$1/16D\times 1/16H\times 1/16W\times 2F$ \\       % & 表示列的分隔线
 \hline
 28 &$3\times 3\times 3\times 64$ stride 1&$1/16D\times 1/16H\times 1/16W\times 2F$ \\       % & 表示列的分隔线
 \hline
 29 &$3\times 3\times 3\times 64$ stride 1&$1/16D\times 1/16H\times 1/16W\times 2F$ \\       % & 表示列的分隔线
 \hline
 30 &$3\times 3\times 3\times 128$ stride 2&$1/32D\times 1/32H\times 1/32W\times 4F$ \\       % & 表示列的分隔线
 \hline
 31 &$3\times 3\times 3\times 128$ stride 1&$1/32D\times 1/32H\times 1/32W\times 4F$ \\       % & 表示列的分隔线
 \hline
 32 &$3\times 3\times 3\times 128$ stride 1&$1/32D\times 1/32H\times 1/32W\times 4F$ \\       % & 表示列的分隔线
 \hline
 33 &$3\times 3\times 3\times 64$ &$1/16D\times 1/16H\times 1/16W\times 2F$ \\       % & 表示列的分隔线
 & upsampling stride 2& add output of layer 29\\
 \hline
 34 &$3\times 3\times 3\times 64$ &$1/8D\times 1/8H\times 1/8W\times 2F$ \\       % & 表示列的分隔线
 & upsampling stride 2& add output of layer 26\\
 \hline
 35 &$3\times 3\times 3\times 64$ &$1/4D\times 1/4H\times 1/4W\times 2F$ \\       % & 表示列的分隔线
 & upsampling stride 2& add output of layer 23\\
 \hline
 36 &$3\times 3\times 3\times 32$ &$1/2D\times 1/2H\times 1/2W\times F$ \\       % & 表示列的分隔线
 & upsampling stride 2& add output of layer 20\\
 \hline 
 37 &$3\times 3\times 3\times 1 $ stride 1 &$D\times H\times W\times 1$ \\ \hline
 \hline
\end{tabular}

\label{tab:cost} 
\end{table}

\subsubsection{A3: Similarity Computation}
The matching cost is designed to compute the similarities of corresponding pixels. The key of cost computation is the similarity measurement between two pixels. As we have obtained the feature volume, we expect to learn a similarity measurement as
\begin{equation}C=T(F),\end{equation}
which is designed to transform the feature volume into a cost volume. Each element of the cost volume represents the similarity computed from the corresponding element of the feature volume. 

3D convolutional networks are effective to take into account the context and geometry information and operate computation from the height, width and disparity three dimensions \cite{kendall2017end}. However, the 3D convolutional operation commonly suffers from the burden on both computational time and intermediate results storage. With the auto-encoder structure, the computational burden can be reduced by subsampling and upsampling operations.

The illustration of the auto-encoder with 3D convolutional layers is presented in Figure \ref{fig:all}.A3 and layer setting is shown in Table \ref{tab:cost}. We apply four sub-sampling units as the encoder and four up-sampling units as the decoder. For the encoder, each sub-sampling unit consists of three 3D-convolution layers and the first convolution layer is applied with $  2\times 2\times 2 $ stride.  For the decoder, the up-sampling unit is realized by one 3D convolution layer with $  2\times 2\times 2 $ stride, besides, the convolution output adds the same resolution feature map from the last layer of the corresponding sub-sampling unit in the encoder.

Since we apply a sub-sampling in feature extraction, to reach the same resolution as the original image, we add an extra up-sampling unit with a single convolution layer. The final output of cost computation is a cost volume with size of $D,H,W$ and each element $C(d,h,w)$ in the volume indicates the matching cost between pixel $R(h,w) $ in the reference image and pixel $T(h,w-d)$ in the target image.

\subsection{Cost Aggregation}
The cost aggregation method is employed to rectify the mismatching cost value computed from the local feature according to the global view guidance. Besides, the cost aggregation can ensure a high-quality disparity map with smoothness and continuity. Through the matching cost computation, we get the initial cost volume $C_0(D, H, W) $. In general, the cost aggregation generates support regions on the cost volume by a statistic or dynamic similarity function. Obtained the regions, the aggregating can be formulated as the convolutional operation on the cost volume, which is expressed as
\begin{equation}C(d,h,w)=W(d,h,w)\otimes C_0 (d,h,w),\end{equation}
where $W$ represents filters and $\otimes$ indicates the convolutional operation.

\begin{algorithm}
\small
\label{aggregation}
\caption{Deep Cost Aggregation}%算法名字
\LinesNumbered %要求显示行号
\KwIn{Initial Cost Volume $C_0(d,h,w)$\\  \qquad  \quad 	Reference Image $I(h,w,3)$}%输入参数
\KwOut{Aggregated Cost Volume $C_a(d,h,w)$}%输出
$\backslash *$  Generation of proposals $* \backslash $\\
Step 1: Aggregation along depth dimension: $C_d(d,h,w,g)=C_0(d,h,w,1)\otimes F_d$ \;
Step 2: Aggregation along height dimension $C_h(d,h,w,g)=C_d(d,h,w,g)\otimes F_h$ \;
Step 3: Aggregation along width dimension $C_w(d,h,w,g)=C_h(d,h,w,g)\otimes F_w$ \;
Step 4: Normalization of aggregation proposals $C_p(d,h,w,g)=C_w(d,h,w,g)\otimes F_0$ \;

$\backslash *$ Extraction of Guidance for Cost Aggregation $* \backslash $\\
Step 5: $G_0(h,w,g)=I(h,w,3)\otimes F_0$\;
Step 6: $G_1(h,w,g)=G_0(h,w,g)\otimes F_1$\;
Step 7: $G_2(h,w,g)=G_1(h,w,g)\otimes F_2$\;
$\backslash *$ Fusion and Selection $* \backslash $\\
Step 8: Fusing the two output from the two-stream netwok: \\$C_f=C_p(d,h,w,g) \odot G_2(h,w,g)$\\
Step 9: Choosing the best evaluated proposal: $C_a(d,h,w)=max\left \{ C_f(d,h,w,g)\right \}$

\end{algorithm}

Compared with the traditional cost aggregation using hand-designed similarity measurement, we propose a learning-based cost aggregation using a two-stream network. The proposed cost aggregation can be directly employed on the initial cost volume and cooperate with the deep cost computation network in an end-to-end trainable manner. Instead of using the deep neural network as a black box, we present an explicit way to leverage the neural network. The cost aggregation is formulated as the selection of cost aggregation proposals, where proposals are potential cost aggregation results. As a result, the two-stream network is designed: one stream for generating the cost aggregation proposals, the other for selecting the best proposals. The proposal stream uses a 3D convolutional network to produce possible cost aggregation results. The results maintain the large receptive field and the rich semantic information transferred from cost computation. The guidance stream directly extracts information from the reference image with a 2D convolutional network. A light convolutional network is employed to extract the low-level structure information as the global view guidance for the selection. 

Many works on understanding deep neural networks \cite{mahendran2015understanding,zeiler2014visualizing} have found that features of the first several convolutional layers are rich in low-level structure information. In contrast, the features from the last several layers have strong high-level semantic information. Both the structure and semantic information is crucial for the cost aggregation. The proposal stream maintains the semantic information, while the guidance stream brings into structure information. The rich semantic information is implicit in the generated proposals, and the structure information is used as global view guidance to evaluate each proposal. The cost aggregation is explicitly carried out by the fusion of these two streams. The details of our two-stream network will be discussed in the following two sub-sections.

\subsubsection{B1: Proposal Sub-network}
The proposal stream is designed to generate the possible cost aggregation results by aggregating matching cost values along the height, width, and depth three dimensions. The aggregating operation is implemented by a 3D convolutional network with rectangle filters. The 3D convolutional network maintains the large view field from the previous cost computation step. The structure of the proposal sub-network is illustrated in Figure \ref{fig:all}.B2. Three 3D convolutional layers are adopted to the initial cost volume. We first use $3\times 1 \times 1$ convolutional filters to aggregate the cost values along the depth dimension, then employ $1\times 3\times 1$ and $1\times 1\times 3$ filters along the height and width dimensions. The rectangle convolutional filters are used to simulate the cost value aggregation process along different dimensions. Compared with the general square filters, the rectangle filter can run in a more explicit manner to aggregate information along different dimensions while actively reduce the computational burden for the 3D convolutional operation. Finally, a convolutional layer with $1\times 1\times 1$ filter is employed to summarize the potential cost aggregation results into $G$ potential aggregation proposals with the size of $D\times H\times W\times G $, where $G$ represents the number of cost aggregation proposals.

The operation along one dimension can be expressed as
\begin{equation}C(d,h,w)=F_i(d,h,w)\otimes C_0 (d,h,w),\end{equation}
where $F$ represents the rectangle filters, $i$ donates the convolutional direction, and $\otimes$ indicates the convolutional operation.

\subsubsection{B2: Guidance Stream}
Since proposals are computed from features of the last layer which has strong semantic information but lacks low-level structure information. The guidance stream is designed to introduce the structure information as the global view guidance to the selection of proposals. It can extract structure information from the reference image to evaluate the generated proposals. As shown in Figure \ref{fig:all}.B1, we employ 2D convolutional network on the reference image to extract the low-level structure information. The convolutional filter is set from $5\times 5$ to $3\times 3$ which can equip the structure information with a large field of view. Moreover, a final $1\times 1$ filter is employed to summarize the guidance to the size of $H\times W\times G $ corresponding to the generated proposals. Furthermore, the guidance is converted into probability value using the softmax method along the dimension of $G$, which ensures that the sum of the evaluation of the proposals is 1. Since we hypothesize the guidance for cost aggregation at different disparities is unchanged, the computed probability value can be treated as the evaluation for different aggregation proposals. The guidance $G_{2}(H,W,i)$ is used as the evaluation for the proposal $C_{a}(D,H,W,i)$. 

In the end, the selection of proposals is achieved by a fusion scheme. The fusion uses the guidance to evaluate the proposals and choose the best evaluation of the fusion results to form the aggregated cost volume. The global view guidance evaluates its corresponding aggregation proposal by a matrix multiplication in a broadcasting manner. The evaluation for each proposal is based on the structure information of the whole reference image so the guidance for the selection is global view. The aggregated cost volume can be obtained by selecting the maximum value along the dimension of $G$. The fusion scheme is indicated as
\begin{equation}C_a(d,h,w)=max\left \{ C_p(d,h,w,g)*C_g(h,w,g) \right \},\end{equation}
where $C_p$ are proposals, $C_g$ represents the guidance, $*$ donates the matrix multiplication and $max$ indicates the maximum function. The process of conducting the cost aggregation algorithm is shown in Algorithm 1.

\subsection{C: Disparity computation}
The aggregated cost volume will be transformed into disparity through a soft-argmin function similar to \cite{kendall2017end} which can retain a sub-pixel disparity accuracy. The matching cost value is converted into probability value by a softmax function along the dimension of depth. The final disparity is obtained by the weighted sum of the probability, where the weights are the corresponding depth value $d$. The mathematical equation is given by
\begin{equation}D(h,w)=\sum _{d=0}^{D_{max}}d\times\sigma (-C_a(d,h,w)),\end{equation}
where $\sigma$ donates the softmax function, $C_a$ is the aggregated cost volume and d is the disparity.

Compared with the traditional WTA strategy, the soft-argmin function can enable the computed disparity influenced by the cost value of all disparity. Therefore, a better smoothness and sub-pixel level accuracy can be obtained. Besides, the soft-argmin is fully differentiable, which ensures that the training can be carried out using back-propagation.

We train the model using the $\ell_1$ error between the ground truth and the predicted disparity. The supervised loss is defined as 
\begin{equation}Loss=\sum_h \sum_w \left \| D_a(h,w)-D_g (h,w)\right \|_1 ,\end{equation}
where $\| \cdot \|_1$ donates the $\ell_1$ norm, $D_g$ is the ground truth disparity map and $D_a$ represents the predicted disparity map.

\section{Experimental Results}
We evaluate our method on three datasets, including Scene Flow \cite{mayer2016large}, KITTI2015 \cite{menze2015object} and KITTI2012 \cite{geiger2012we}. We especially compare our method with the state-of-the-art GC-Net \cite{kendall2017end} to demonstrate the effectiveness of the learning-based cost aggregation. Our architecture is implemented by the Tensoflow \cite{abadi2016tensorflow} with a standard RMSProp \cite{tieleman2012lecture} and a constant learning rate of 0.0001. We train the network on the Scene Flow dataset from a random initialization with shuffled orders. The training takes 23h after 300K iterations on a single NVIDIA 1080Ti GPU. For the KITTI dataset, we fine-tune the model pre-trained on Scene Flow dataset with 70k iterations. Limited by the computation resource, we sub-sample all data by four times using the bilinear interpolation. 

\begin{table}[tbp]
\caption{Comparisons on Scene Flow}
\centering  % 表居中
\scriptsize
\begin{tabular}{l|c|c|c|c}  % {lccc} 表示各列元素对齐方式,left-l,right-r,center-c
\hline
Model &error $>$ 1px &error $>$ 3 px &MAE(px) &T(ms)\\ \hline \hline % \hline 在此行下面画一横线
GC-Net &11.3 & 7.2&2.21&0.95\\     \hline   % & 表示列的分隔线

Without guidance&12.3 &7.2&2.15&0.93\\         % \\ 表示重新开始一行
Without proposal&10.81 &6.8&1.83&0.85\\         % \\ 表示重新开始一行
Without aggregation&13.8 &7.5&2.71&0.95\\         % \\ 表示重新开始一行
\textbf{Our model}&\textbf{8.93}&\textbf{5.62}&\textbf{1.75}&\textbf{1.12}\\         % \\ 表示重新开始一行
 \hline
\end{tabular}

\label{scene}
\end{table}

\begin{table}[tbp]
\tiny
\caption{Comparisons on KITTI2012}
\centering  % 表居中
\begin{tabular}{l|c|c|c|c|c|c|c}  % {lccc} 表示各列元素对齐方式,left-l,right-r,center-c
\hline
Model &\multicolumn{2}{|c|}{$>$2px } &\multicolumn{2}{|c|}{$>5$ px}&\multicolumn{2}{|c|}{Mean Error }&T(s)\\ 
 &Non-Occ&All&Non-Occ&All&Non-Occ&All&\\ 
\hline \hline % \hline 在此行下面画一横线
PSMNet&\textbf{ 2.62} & \textbf{3.24} & \textbf{0.94}  & \textbf{1.20}&\textbf{0.5}&\textbf{0.6}&1.3\\
GC-Net& 2.71 & 3.46 & 1.77  & 2.30&0.6&0.7&0.9\\
SegStereo& 3.24 & 3.82 & 1.10  & 1.35&0.6&0.6&0.6\\       % & 表示列的分隔线
Displets v2&3.43 & 4.46 &1.72  & \textbf{2.17}&0.7&0.8&265\\         % \\ 表示重新开始一行    
L-ResMatch &3.64  & 5.06 &\textbf{1.50 } & 2.26&0.7&1.0&48\\       % & 表示列的分隔线
MC-CNN &3.90  & 5.45 &1.64  & 2.39&0.7&0.9&67\\         % \\ 表示重新开始一行  
CATN &8.11  & 9.44 &3.31  & 4.07&1.1&1.2&10\\         % \\ 表示重新开始一行 
S+GF &14.72  & 16.76 &5.53  & 7.79&2.1&3.4&140\\         % \\ 表示重新开始一行 
\hline   % & 表示列的分隔线
Our model&2.68 &3.42 &1.63&2.23&0.6&0.7&1.13\\        % \\ 表示重新开始一行
\hline 
\end{tabular} 

\label{kitti2012}
\end{table}
\begin{table}[tbp]
\centering  % 表居中
\caption{Comparisons on KITTI2015}
\tiny
\begin{tabular}{l|c|c|c|c|c|c|c}  % {lccc} 表示各列元素对齐方式,left-l,right-r,center-c
\hline
Model &\multicolumn{3}{|c|}{All pixels} &\multicolumn{3}{|c|}{Non-Occluded Pixels}&Time(s)\\ 
 &D1-bg& D1-fg& D1-all& D1-bg& D1-fg &D1-all&\\ 
\hline \hline % \hline 在此行下面画一横线      % & 表示列的分隔线
PSMNet &\textbf{1.97 }&4.41 &\textbf{2.38}&\textbf{1.81}&4.00&\textbf{2.17}&1.3\\
SegStereo &2.16 &\textbf{4.02} &2.47&2.01&\textbf{3.62}&2.28&\textbf{0.6}\\    
GC-Net &2.21 &6.16 &2.87&2.02&5.58&2.61&0.9\\       % & 表示列的分隔线
MC-CNN&2.89 &8.88 &3.89&2.48&7.64&3.33&67\\         % \\ 表示重新开始一行
Displetv v2&3.00 &5.56 &3.43&2.73&4.95&3.09&265\\ 
DRR&2.58 &6.04 &3.16&2.34&4.87&2.76&0.4\\ 
L-ResMatch&2.72 &6.95 &3.42&2.35&5.74&2.91&48\\ 
3DMST&3.36 &13.03 &4.97&3.03&12.11&4.53&93\\ 
\hline   % & 表示列的分隔线
Our model&2.17 &5.46 &2.79&2.06&5.32&2.32&1.13\\        % \\ 表示重新开始一行
\hline 
\end{tabular} 

\label{kitti2015}
\end{table}

\subsection{Benchmark results}

Scene Flow is a synthetic data set for stereo matching which contains $35454$ training and $4370$ testing image pairs. Synthetic dataset ensures dense ground truth without inaccurate labels and is large enough to train a complex network without over-fitting. In Table \ref{scene}, we evaluate our method and GC-Net on the Scene Flow dataset. We observe that our method outperforms GC-Net among all pixel errors and the RMS error. In addition, to demonstrate the effectiveness of each stream of our network, we evaluate the network with different settings. From Table \ref{scene}, we can see the guidance stream is crucial to improving the performance, which demonstrates the structure information can be used as global view guidance to improve the accuracy.

The KITTI benchmark consists of challenging and complex road scene collected from a moving vehicle. The ground truth of disparity image is obtained from LIDAR data. The KITTI 2012 dataset contains $192$ training and $195$ testing images, and the KITTI 2015 dataset contains $200$ training and $200$ testing images. In the Table \ref{kitti2012}, the comparisons on KITTI2012 with deep stereo methods such as GC-net \cite{kendall2017end}, Displets v2 \cite{guney2015displets}, L-ResMatch \cite{shaked2016improved} and MC-CNN \cite{zbontar2015computing} are shown, besides, the comparisons with other cost aggregation methods including CAT \cite{ha2014cost} and S+GF \cite{zhang2014cross} are also illustrated in Table \ref{kitti2015}, the leaderboard on KITTI2015 compares our method with GC-Net \cite{kendall2017end}, MC-CNN \cite{zbontar2016stereo}, Displetv v2 \cite{guney2015displets}, DRR \cite{gidaris2016detect}, L-ResMatch \cite{shaked2016improved} and 3DMST \cite{li20173d}. Our method outperforms previous works which use a hand-designed aggregation method or ignoring the aggregation step. It can be inferred that the usage of learning-based cost aggregation method can improve the performance of the deep stereo matching.

\subsection{Sub-architecture Analysis}
\begin{figure}
\centering  
\includegraphics[width=8.3cm]{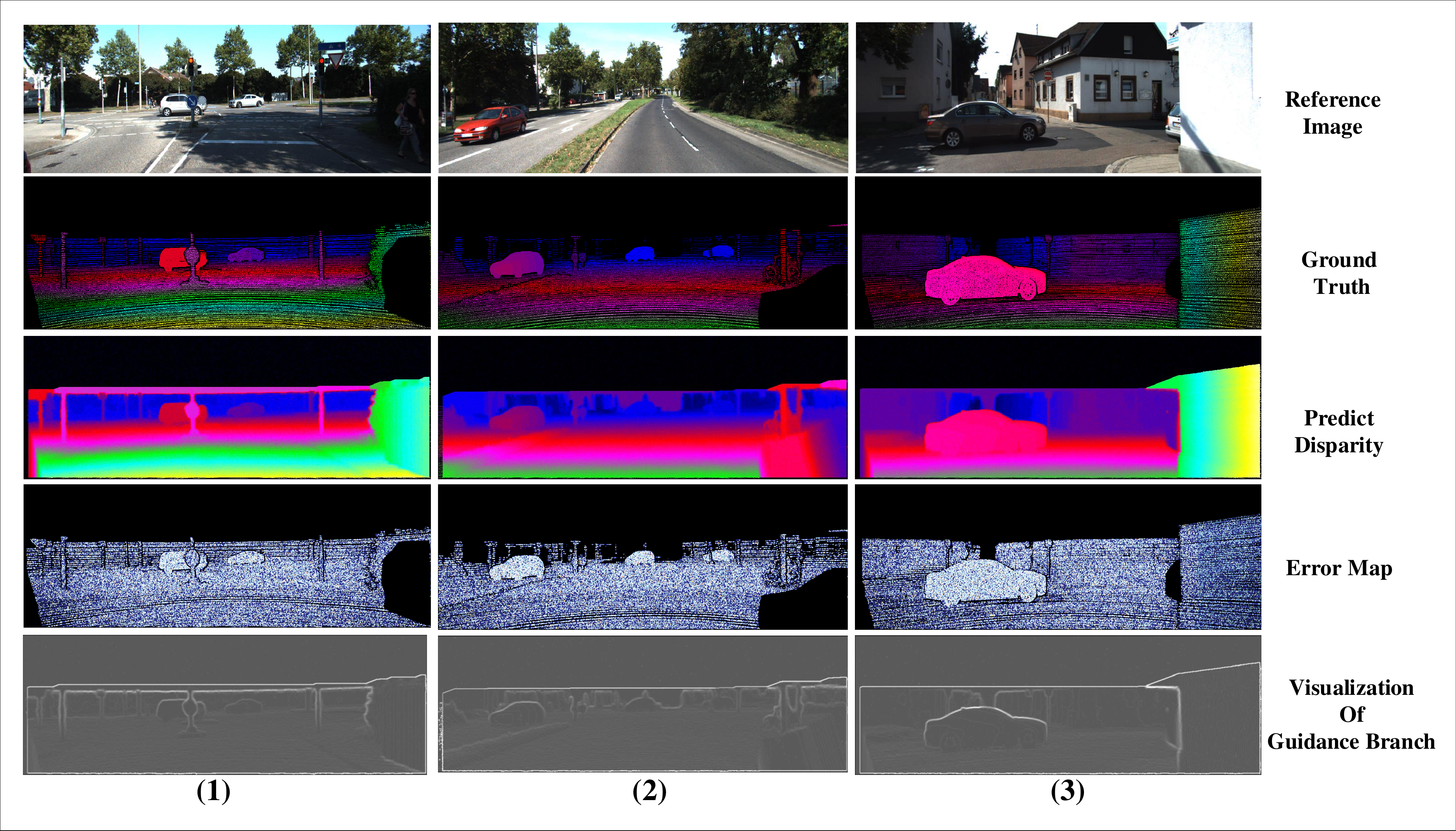}
\caption{The visualization of our experimental results. From top to bottom, images are the reference image, ground-truth disparity map, predicted disparity map, error map and the visualization of the output from our guidance stream, respectively. The visualization of the guidance stream shows that it exactly extracts structure information.
}  
\label{fig:visual} 
\end{figure}

To demonstrate the effectiveness of the learning-based cost aggregation, we visualize the guidance obtained from the guidance stream in this section. According to the visualization of the Figure \ref{fig:visual}, we can infer that the guidance stream can obtain the structure information from reference image which can select the aggregation proposal with a global view. The visualized feature map of guidance sub-network is realized by averaging the output of the guidance stream along the dimension $G$. We can obviously see the guidance contains low-level structure information, which demonstrates that the two-stream network can introduce structure information as the global view guidance for the selection of proposals.

\section{Conclusion}
In this paper, we have proposed a learning-based cost aggregation for stereo matching. The learning-based cost aggregation can be embedded into the deep stereo matching solution in an end-to-end manner. With this end-to-end trainable manner, our cost aggregation achieved a higher accuracy by effectively collaborating with the deep cost computation methods. According to the analysis of the two-stream network, we demonstrated that the low-level structure information can be used as global view guidance for selection of the proposals of the rich semantic information. Furthermore, the proposed two-stream network had the potential ability for feature fusion works such as motion recognition and scene understanding. The experiment results have demonstrated the good ability of our explicit architecture for stereo matching.
\section{Acknowledgement}
This work was supported in part by the Natural Science Foundation of China (NSFC) under Grants No. 61773062 and No. 61702037.
\bibliographystyle{aaai}
\bibliography{reference}
\end{document}